# Medical Image Retrieval using Deep Convolutional Neural Network

[1]Adnan Qayyum, [2a*]Syed Muhammad Anwar, [3b]Muhammad Awais, [1a]Muhammad Majid

[1]Department of Computer Engineering, University of Engineering and Technology Taxila, Pakistan

[2]Department of Software Engineering, University of Engineering and Technology Taxila, Pakistan

[3]Center for Vision, Speech and Signal Processing, University of Surrey, Surrey, UK

[a]Signal, Image, Multimedia Processing and Learning (SIMPLe) Group, University of Engineering and Technology, Taxila, Pakistan

[b]Cerebrai Artificial Intelligence, Surrey, UK

*s.anwar@uettaxila.edu.pk

**Abstract**

With a widespread use of digital imaging data in hospitals, the size of medical image repositories is increasing rapidly. This causes difficulty in managing and querying these large databases leading to the need of content based medical image retrieval (CBMIR) systems. A major challenge in CBMIR systems is the semantic gap that exists between the low level visual information captured by imaging devices and high level semantic information perceived by human. The efficacy of such systems is more crucial in terms of feature representations that can characterize the high-level information completely. In this paper, we propose a framework of deep learning for CBMIR system by using deep Convolutional Neural Network (CNN) that is trained for classification of medical images. An intermodal dataset that contains twenty four classes and five modalities is used to train the network. The learned features and the classification results are used to retrieve medical images. For retrieval, best results are achieved when class based predictions are used. An average classification accuracy of 99.77% and a mean average precision of 0.69 is achieved for retrieval task. The proposed method is best suited to retrieve multimodal medical images for different body organs.

## 1. Introduction

In recent years, rapid growth of digital computers, multimedia, and storage systems has resulted in large image and multimedia content repositories. Clinical and diagnostic studies are also benefiting from these advances in digital storage and content processing. The hospitals having diagnostic and investigative imaging facilities are producing large amount of imaging data thus causing a huge increase in production of medical image collections. Therefore, development of an effective medical image retrieval system is required to aid clinicians in browsing these large datasets. To facilitate the process of production and management of such large medical image databases, many algorithms for automatic analysis of medical images have been proposed in literature [1-5]. A content based medical image retrieval (CBMIR) system can be an effective way for supplementing the diagnosis and treatment of various diseases and also an efficient management tool [6] for handling large amount of data.

Content based image retrieval (CBIR) is a computer vision technique that gives a way for searching relevant images in large databases. This search is based on the image features like color, texture and shape or any other features being derived from the image itself. The performance of a CBIR system mainly depends on these selected features [7]. The images are first represented in terms of features in a high dimensional feature space. Then, the similarity among images stored in the database and that of a query image is measured in the feature space by using a distance metric e.g., Euclidean distance. Hence, for CBIR systems, representation of image data in terms of features and selecting a similarity measure, are the most critical components. Although, many researchers have broadly studied these areas [8], but the most challenging issue that remains in CBIR systems is reducing the "semantic gap". It is the information lost by representing an image in terms of its features i.e., from high level semantics to low level features [9]. This gap exists between the visual information captured by the imaging device and the visual information perceived by the human vision system (HVS). This gap can be reduced either by embedding domain specific knowledge or by using some machine learning technique to develop intelligent systems that can be trained to act like HVS.

There has been a significant growth in machine learning research and one breakthrough is deep learning framework. The deep learning possesses various machine learning algorithms for modelling high level abstractions in data by employing deep architectures composed of multiple non-linear transformations [10]. Deep learning mimics the human brain [8], that has a deep architecture and information in human brain is processed through multiple layers of transformation. Thus, to learn features from data automatically at multiple level of abstractions by exploring deep architectures, deep learning techniques gives a direct way to get feature representations by allowing the system (deep network) to learn complex features from raw images without using hand crafted features. Recent studies have reported that deep learning methods have been successfully applied to many applications areas e.g., image and video classification [11-13], visual tracking [14], speech recognition [15] and natural language processing [16].

Deep learning methods have been applied to CBIR task in recent studies [8, 17, 18], but there is less attention on exploring deep learning methods for CBMIR task. In this paper, inspired by the successes of deep learning methods in bridging the semantic gap, its application to CBMIR task is investigated. A deep learning technique i.e., Convolutional Neural Network (CNN) is adapted for learning feature representations for different imaging modalities and body organs. Generally, in medical imaging 3D volumetric image are obtained consisting of a series of 2D slices acquired from the target body organ. This paper focuses on retrieval of these 2D slices, the classes were formulated at global level i.e., images from different body parts were divided into separate classes with respective body part label. In this way, the supervision is very weak and requires very less time for labelling, hence decreasing the annotation effort required in training phase. For medical imaging, this type of annotation is quite useful since, annotations usually require expert advice and high cost. The CNN model is trained for classifying medical images in first phase and then the learned feature representations are used for CBMIR in second phase. An in-depth analysis of the proposed system in terms of retrieval quality is presented for a collection of images belonging to different imaging modalities. The major contributions of this work are threefold,

I. A dataset is carefully collected that is multimodal and covers a wide range of medical imaging target areas.
II. A deep learning framework is modelled and trained on a collection of medical images.
III. The learned features are used to present a highly efficient medical image retrieval system that works for a large collection of multimodal dataset.

The rest of the paper is organized as follows. Related work is presented in Section 2, the proposed methodology is discussed in Section 3, experimental results are shown and discussed in Section 4 and a conclusion is presented in Section 5.

## 2. Related Work

In this section, existing work related to our research is briefly discussed.

### 2.1 Content Based Image Retrieval (CBIR)

A basic block diagram of generic CBIR system is illustrated in Fig. 1. In CBIR, the images are retrieved from large databases, based on feature representations extracted or derived from the image content [9]. There are typically two phases in any CBIR system, first one is offline and the other is online phase. In the offline phase, features are extracted from large collections of images (used to train the system) to establish a local features database. This phase is generally time consuming and depends on the number of training images used to train the system. In the online phase, same features are extracted from the query image and a distance metric is calculated between the features of query image and features of database images for similarity measure. Those images having high similarity or low distance are then presented to the user as retrieval results. The procedure used for pre-processing and feature extraction is same in both phases.

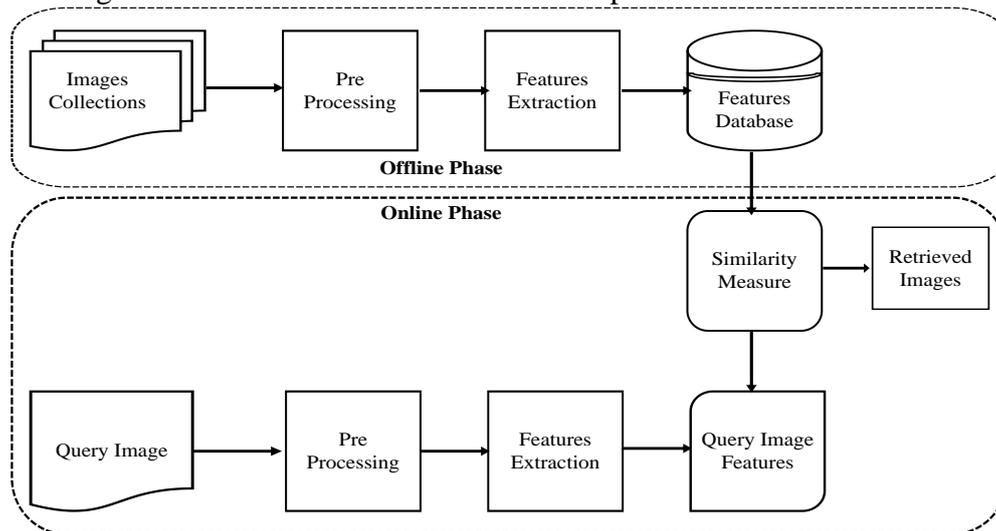

Fig. 1. Block diagram of a generic CBIR system.

In the past decades, various feature descriptors have been developed for representing images at a global level such as shape and color based features [19], texture based features [20]. Local level feature descriptors have also been developed such as Scale Invariant Feature Transform (SIFT) [21], Speeded up Robust Features (SURF) [22] and Bag of Words (BoW) [23-25] model using some local descriptors like SIFT [21] or SURF [26]. These features have been used in various studies for image retrieval task [27-29] but are still not fully able to address the problem of semantic gap. Recently, machine learning techniques have been explored to address the problem of semantic gap using local-level features, a few of them focused on learning hash or compact codes [30-32]. The advanced machine learning techniques have provided a new way of reducing the semantic gap and deep learning has given a hope for bridging this gap by learning visual features directly from the images without using any hand-crafted features.

## 2.2 Content Based Medical Image Retrieval (CBMIR)

With the widespread dissemination of Picture Archiving and Communication Systems (PACS) in hospitals, the size of medical image collections are increasing rapidly [33]. Therefore, to manage such large medical databases, development of effective medical image retrieval system is required. Apart from this task of managing database, a specific CBMIR system helps the doctors in making critical decisions about a specific disease or injury. By retrieving similar images and case histories, the doctors could make a more informed decision about the patient's disease stage and diagnosis [1]. For medical images, global feature extraction based systems have failed to provide compact feature representations as clinically beneficial information is highly localized in small regions of the image [9].

In [1], an approach based on Bag of Visual Words (BoVWs) using SIFT features was presented for brain magnetic resonance image (MRI) retrieval for diagnosis of Alzheimer disease. They proposed Laguerre Circular Harmonic Functions coefficients (LG-CHF) as feature vectors for image matching. A method for content based retrieval for skin lesion images using reduced feature vector, classification and regression tree was presented in [2]. In [3] impact of the result of a medical CBIR system were proposed on the decision of doctor is studied using a CBIR algorithm for mammographic images by using different features extraction techniques, various distance measures and relevance feedback. In [4], two CBIR methods based on Wavelets adaptation, where a different Wavelet basis was used for characterizing each query image and to estimate the best Wavelet filter. A regression function that was tuned for maximizing retrieval performance was used. In [5], a classification driven supervised learning approach towards biomedical image retrieval was proposed. It has used image filtering and similarity fusion as basis and multiclass support vector machine (SVM) was used to predict the class of query image. Hence, by eliminating the irrelevant images the search area is reduced in large database for similarity measurement.

## 2.3 Deep Learning

Deep learning is a subfield of machine learning, which uses set of algorithms that attempts to model high level abstractions present in data by using a deep architecture possessing multiple processing layers, having linear and nonlinear transformation functions [34]. The history of deep learning started in 1965 [35] but has only seen major advances recently with the availability of improved computational capabilities, nonlinearities which allow for deeper network [36,37] and better ways to initialize deep network [38]. Deep learning is based on artificial neural networks which attempts to mimic the way human brain works. The feed forward neural networks, comprising of many hidden layers are good example of models with deep architecture. The standard back propagation algorithm popularized in 1980 is still an effective way of training the neural networks [8]. A standard back propagation algorithm for training neural networks with multiple layers was used for handwritten digit recognition [39]. A Deep Convolutional Neural Network (DCNN) presented in [11] has won the image classification challenge of ILSVRC-2012 triggering exponential growth in the market.

Recent studies have shown that deep learning techniques are successfully applied to medical domain. In [40], convolutional neural network based system was presented for classification of Interstitial Lung Diseases (ILDs). Their dataset comprised of 7 classes, out of which 6 were ILD patterns and a healthy tissue class. They achieved a classification performance of 85.5% in characterizing lungs patterns. In [41] a convolutional classification restricted Boltzmann machine based approach was presented for lung computed tomography (CT) image analysis that combines generative and discriminative representation learning. They presented two approaches for two different datasets: one for lung texture classification and other for airway detection. In [42], multi scale CNN was applied for automatic segmentation of MR images by classifying voxel into brain tissue classes. The network was trained on multiple sized image patches with different kernel size depending on the patch size being used. In [43], a two-stage multiple instance deep learning framework was presented for body organ recognition. In the first stage, the CNN was trained on local patches to extract discriminative and non-informative patches from training samples. In second stage, the network was fine-tuned on extracted discriminative patches for classification task, the dataset consisted of 12 classes having CT and MR 2D slices. The application of DCNN in computer aided diagnosis (CAD) was presented in [44]. Three aspects of CNN were studied i.e., different CNN architectures, dataset scale and transfer learning, where a pre-trained model trained on general images was fine-tuned for medical domain.

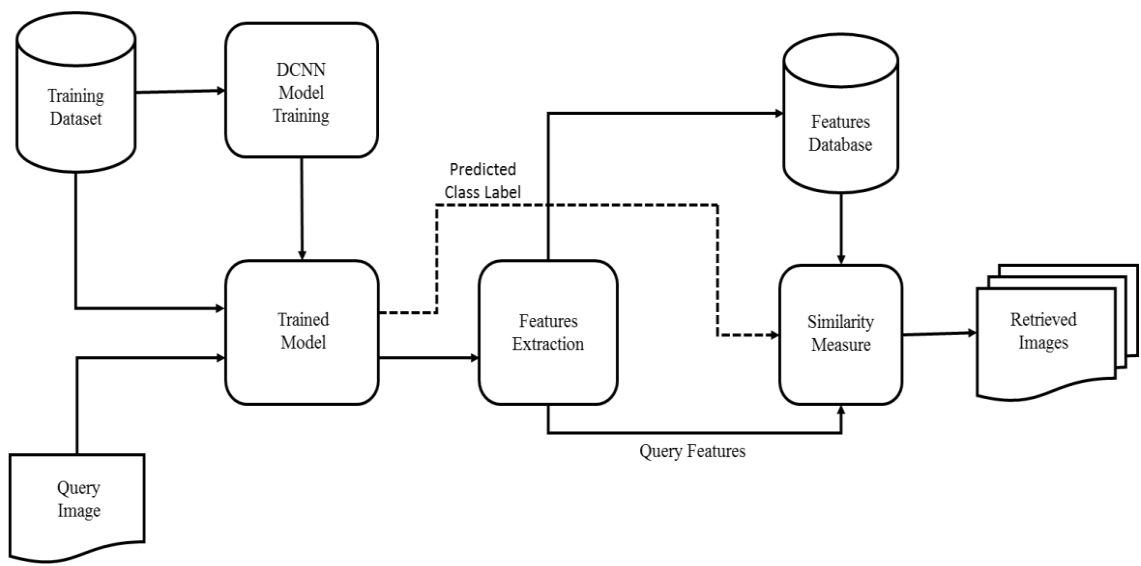

Fig. 2. The proposed framework for content based medical image retrieval using deep convolutional neural network

## 3. Methodology

In this work, a classification driven framework for retrieving similar images from medical database is proposed. A detailed representation of the proposed retrieval system is shown in Fig. 2. The underlying DCNN model aims to learn filter kernel by producing a more abstract representation of the data in each layer. Despite of its simple mathematics, DCNN is currently the most powerful tool in vision systems. The DCNN models generally have three types of layers i.e., convolutional layers, pooling layers, and fully connected layers. The output layer is generally treated separately as a special layer and the model gets input samples at the input layer. Each convolutional layer produces feature maps by convolving the kernel with input feature maps. A pooling layer is designed to down sample feature maps produced by the convolutional layers, which is often accomplished by finding local maxima in a neighborhood. Also, pooling gives translational invariance and in the meanwhile it reduces the number of neurons to be processed in upcoming layers. In fully connected layers, each neuron has a more denser connection as compared to the convolutional layers. The part of the DCNN before fully connected layers is known as feature extractor part and after that is known as classifier part. A detailed description of the framework used is presented in following subsections.

### 3.1 Phase 1: Classification

The first phase in our proposed framework of CBMIR is the classification phase, in which a deep convolutional neural network is trained for classifying medical images by following supervised learning approach. For this purpose, the medical images are divided into various classes based on body part or organ information, more details on the dataset can be found in Section 3. 2D images have been used for analysis, hence the task is to classify each image into a class, which ultimately formulates into a multiclass image classification problem. Typically, image classification algorithms have two modules i.e., feature extraction and classification module. DCNN learns both hierarchy of deep convolutional features and classifier from the training image data in an end-to-end learning framework. Deep learning algorithm learns low-level, mid-level and abstract features directly from the images as opposed to making domain specific assumptions, which is the case for handcrafted features. Hence, it can identify the class of a query image more effectively and therefore the learned features can be used for image retrieval task. Inspired by this property, a DCNN model is trained and optimized for multiclass classification problem. The learned features are extracted from the trained model for retrieval. The details about the model architecture and training are as follows.

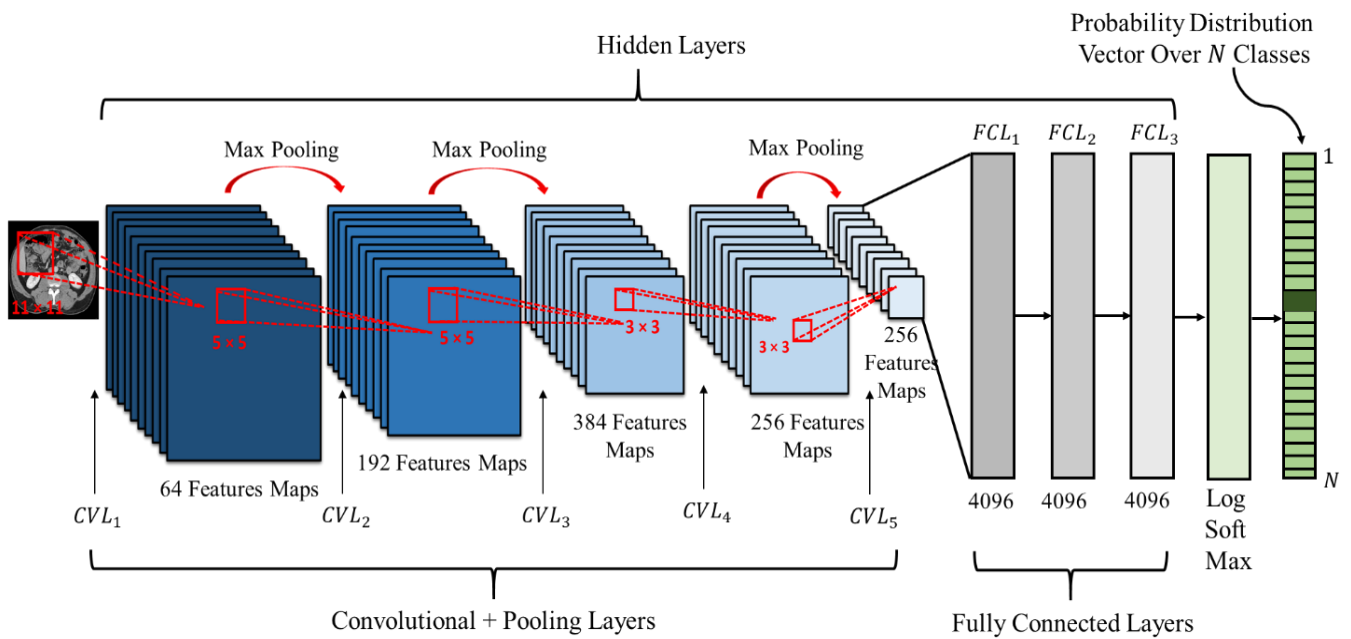

Fig. 3. The DCNN architecture used for the CBMIR task.

### 3.1.1 The DCNN Model Architecture

The model used for training consisted of eight layers, out of which five were convolutional layers and three were fully connected layers, as depicted in Fig. 3. The convolutional and fully connected layers are represented as CVL and FCL, where the subscript represents the layer number e.g., $CVL_1$ represents the first convolutional layer. The output of last fully connected layer ($FCL_3$) has been fed to a softmax function having 24 outputs, which produce a probability distributions for each class label. Hence, probabilities vector of size $1 \times 24$ where each vector element corresponds to a class of dataset is obtained. The network accepts grayscale images of dimension $224 \times 224$ as inputs and unlike the model presented in [11] uses lesser number of kernels. The $CVL_1$ filters the input image with 64 kernels of size $11 \times 11$ with stride equal to 4 pixels. Stride is the distance between the centers of receptive fields of neighborhood neurons in the kernel map. The output of first convolutional layer is fed to a non-linearity and then passed through the spatial max pooling layer for summarizing neighboring neurons. Rectified linear unit (ReLU) [36] nonlinearity is applied to the outputs of all convolutional and fully connected layers. This network with ReLUs has the ability to get trained several times faster than its equivalent with tanh units [37] as well as it also allows to go deeper with vanishing gradient problems. The $CVL_2$ takes the output of $CVL_1$ as input processed by ReLU non-linearity and spatial max pooling layers respectively and filters it with 192 kernels of size $5 \times 5$. The $CVL_3$ contains 384 kernels of size $5 \times 5$ and it gets input from the pooled outputs of the second convolutional layers. Both the convolutional layers, $CVL_4$ and $CVL_5$ have 256 kernels of size $3 \times 3$. All fully connected layers have equal number of neurons i.e., 4096.

After the first, second and fifth convolutional layer, a pooling layer has been used. It consists of a grid of pooling units spaced at 2 pixels apart, each one summarizes the neighboring neurons in neighborhood of $3 \times 3$ that are centered at location of the pooling unit. The overlapping pooling operation makes it difficult for the model to overfit during training. After first and second fully connected layers' dropout regularization layer [45] has been used to avoid further overfitting with a neuron dropout probability of 1-p, where p is the probability of neurons kept. The dropped-out neurons do not contribute to the forward pass as well as in backward pass, i.e., all the connections going into and coming out of dropped-out neurons are removed in a training stage. After the training stage is finished the dropped-out neurons are reinstated with their original weights for next training stage. In the test phase, all the neurons are used without any of them being dropped-out. However, to balance the expected values of the neurons in the test phase to that of the training phase a factor of kept probability 'p' weights them. Finally, the output of $FCL_3$ is fed to input of log softmax with 4096 inputs and 24 outputs. Log softmax applies a soft max function to a N-dimensional vector, which scales the values of the vector in the range of $[0, 1]$ and its summation gives a value of 1. By doing this it assigns probability distribution to each class.

### 3.1.2 Training Details

All images in the database were center cropped using dimensions of $224 \times 224$ prior to training. The model was trained using Stochastic Gradient Descent (SGD) with backpropagation. It was optimized with a very low learning rate of 0.0001 with maximum of 30 epochs of SGD. The Negative Log Likelihood (NLL) was used as an objective function or criterion in SGD training. The NLL is frequently used in classification problems having $N$ classes. The weights of all layers were initialized using Gaussian distribution having mean of zero and standard deviation of 0.01. The biases in the $CVL_2$, $CVL_4$, $CVL_5$ and all three fully connected layers were initialized with constant '1' and rest of the layers with '0'. The learning rate was kept constant for all iterations of stochastic gradient i.e., 0.0001 as for dataset used, decaying the learning rate, increases the training time as compared to keeping it constant. Another reason for keeping learning rate constant is that our dataset was already getting trained by a very low learning rate so reducing it further will require more processing power and obviously, it will take more time. With a maximum of 30 epochs of SGD, training the model was optimized with a training error of 0.0422.

SGD is the most commonly used algorithm for training neural networks and it is very efficient in learning discriminative linear classifiers under a convex loss function like SVM or Logistic Regression. The two major advantages of using SGD are efficiency and ease in implementation providing options in tuning the network like number of iterations, learning rate, rate decay etc. A few disadvantages of SGD include its need of hyper-parameters like number of epochs or iterations and regularization parameters. SGD performs parameters update for each training sample $x^i$ and label $y^j$. Eq. 1 is used for parameter update of the SGD.

$$\theta = \theta - \eta . \nabla_\theta J(\theta; x^i, y^j), \qquad (1)$$

where '$J$' is the objective function that SGD will optimize (Negative Log-Likelihood in our case) and $\theta$ denotes DCNN model parameters i.e., weights and biases.

Backpropagation is the most frequent algorithm used for training artificial neural networks, that is coupled with some optimization technique e.g., SGD. In back propagation, gradients are calculated with respect to all parameters. The SGD gets calculated gradients as input and updates them while trying to minimize the objective function. To compute gradients of loss, backpropagation requires the known target of each input i.e., actual class label. The chain rule is used in an iterative manner for computation of gradients for each layer with respect to loss function. There are three main phases of backpropagation in each layer i.e., forward pass, backward pass, and derivative of the objective function with respect to layer's parameters, if the layer is supposed to have parameters e.g., convolutional layer. Some layers such as pooling layer do not have any parameters and hence the derivative does not need to be computed.

*Forward Pass*
In forward pass, forward message is sent to compute all $z's$, where $z$ is function of input $x$, Eq. 2 gives it mathematical notation.

$$z^{l+1} = f(z^l), \qquad (2)$$

where $l$ is the layer number, and $z = f(x_i)$.

*Backward Pass*
In backward pass, backward message is sent to compute all $\delta's$, where $\delta$ is the derivative of cost function w.r.t $z's$, mathematically it is given as

$$\delta_i^l = \frac{dE}{dz_i^l}$$
$$= \sum_j \frac{dE}{dz_j^l} \cdot \frac{dz_j^{l+1}}{dz_i^l} = \sum_j \delta_j^{l+1} \left(\frac{dz_j^{l+1}}{dz_i^l}\right), \qquad (3)$$

where $i$ is the unit's index for layers, $j$ represents input sample, and $E$ denotes loss function. Eq. 3 is recursive and hence backpropagation with SGD attempts to minimize the loss function recursively.

*Derivative with respect to parameters*
If the layer has parameters, derivatives of cost function w.r.t parameters are computed in backpropagation. Eq. 4 demonstrates a mathematical way of doing this.

$$\frac{dE}{d\theta^l} = \sum_j \frac{dE}{dz_j^{l+1}} \cdot \frac{dz_j^{l+1}}{d\theta^l}$$
$$= \sum_j \delta_j^{l+1} \left(\frac{dz_j^{l+1}}{d\theta^l}\right), \qquad (4)$$

where $\theta$ represents the layer parameters (weights and biases), the rest of variables are same as already described above. Fig. 4 illustrates backpropagation for a neural network with $L$ layers, which are connected in sequential structure. Each of the layer is taking an input and generating an output, it takes loss value $E$ from output layer (loss layer) and back propagates it. The forward messages ($z's$) and backward messages ($\delta's$) are generated by using Eq. 2 and Eq. 3. The derivatives w.r.t parameters ($\theta$) are calculated by using Eq. 4. There can be many such layers in each layer and $\theta$ represents the parameters vector that contains all the parameters from the layer within that specific layer. The gradient of loss w.r.t to layers parameters $\theta$ is given by $\frac{dE}{d\theta}$.

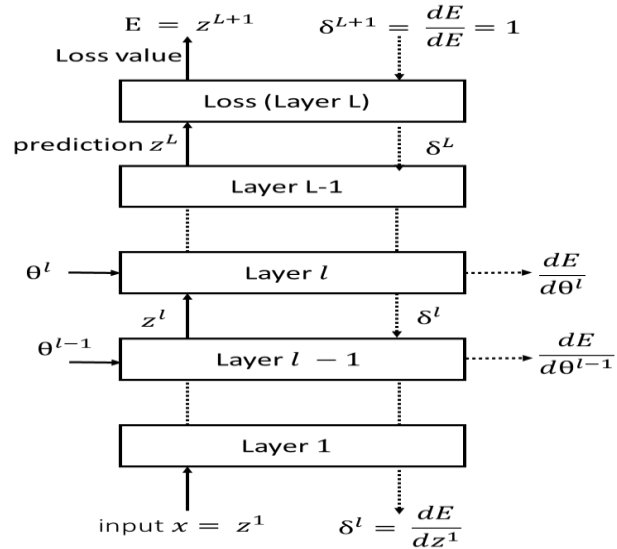

Fig. 4. An illustration of the backpropagation algorithm showing different layers and parameters

### 3.2 Phase 2: Features Extraction for CBMIR

Once the DCNN model is successfully optimized and trained for classifying the multimodal medical images, features representations are extracted from last three fully connected layers of the trained model i.e. from $FCL_1$ - $FCL_3$. For image retrieval task a locally established features database for the whole training data is required. Therefore, to create such features database, each image $x_i$ from training set is feed forwarded to the trained DCNN model for classification task and then features representation $F_{1i}$, $F_{2i}$, and $F_{3i}$ are extracted associated to that specific image from fully connected layers 1 to 3 respectively. The $F_{1i}$, represents a features database extracted from $FCL_1$ and similarly $F_{2i}$ and $F_{3i}$ represents features databases extracted from $FCL_2$ and $FCL_3$, where $i = 1$ to $P$ and $P$ is equal to number of samples in training set. Whenever a query is formulated, similar images as that of query image are retrieved by comparing feature representations extracted for query image (by passing query image from same trained model) and that of features database by using Euclidian distance metric and is given as

$$d(a,b) = \sqrt{\sum_{i=1}^{P}(a_i - b_i)^2} \quad , \qquad (5)$$

where $a_i$ and $b_i$ represent the query and database image features respectively. In addition, the predicted class label (dashed line in Fig. 2) has been used to limit the search area in the database by reducing the number of computations and eliminating irrelevant images from retrieval results. Those images having low distance or high similarity as compared to others are displayed as retrieval results to the user. Finally, comparative analysis is performed for features representations extracted from $FCL_1$ to $FCL_3$ in terms of retrieval quality.

### 4. Experimental Results

In this paper, a popular and widely-used deep learning tool Torch7 [30] has been used for developing and training the proposed deep learning framework. The simulations have been performed on Dell Inspiron 5520 Laptop with Ubuntu 14.04 having Intel Core i3 CPU with clock speed of 2.40 GHz with a RAM of 6.00 GB. The proposed method has been evaluated in terms of classification and retrieval results.

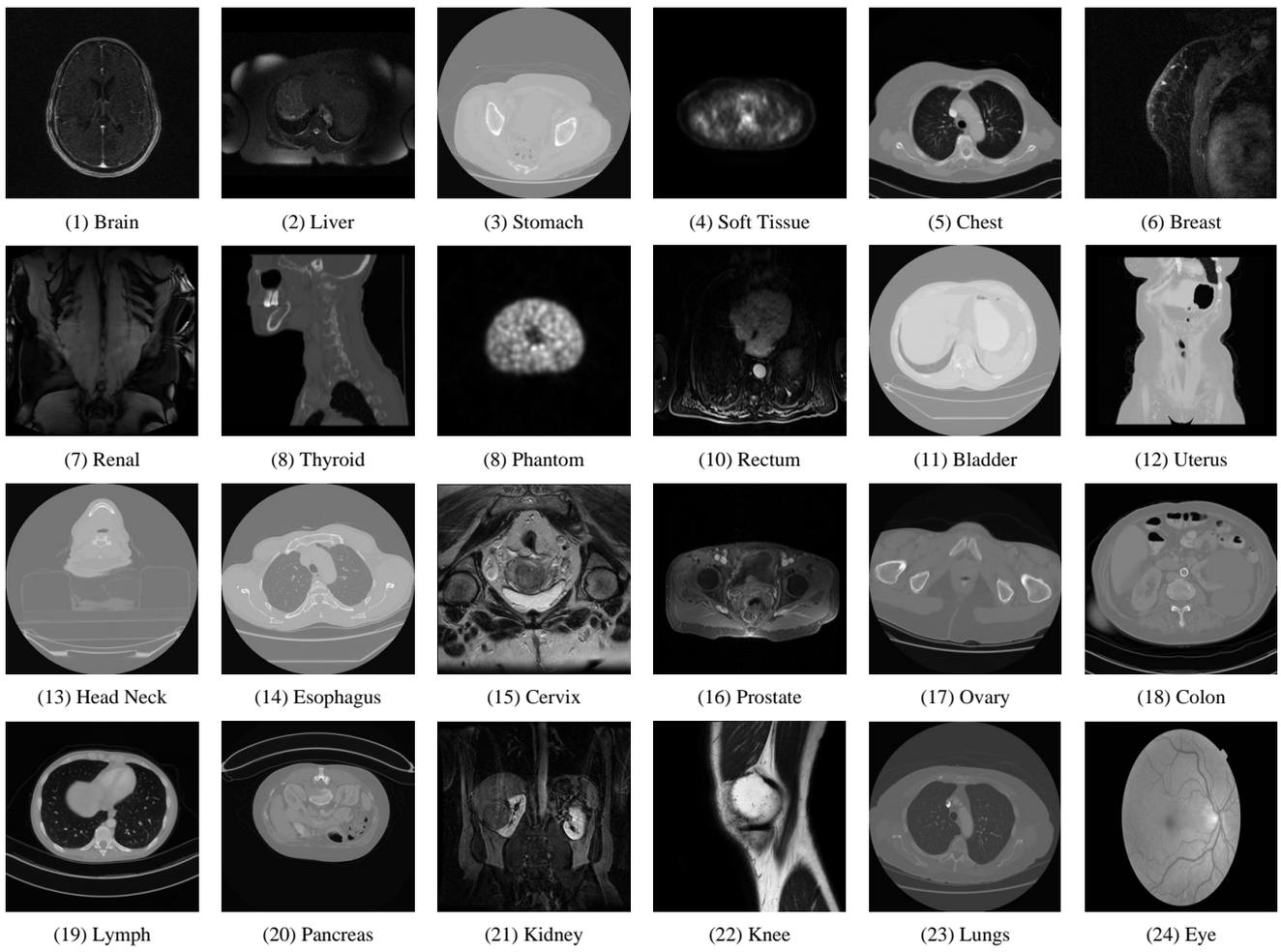

Fig. 5 Example images from each class in the dataset showing interclass variations

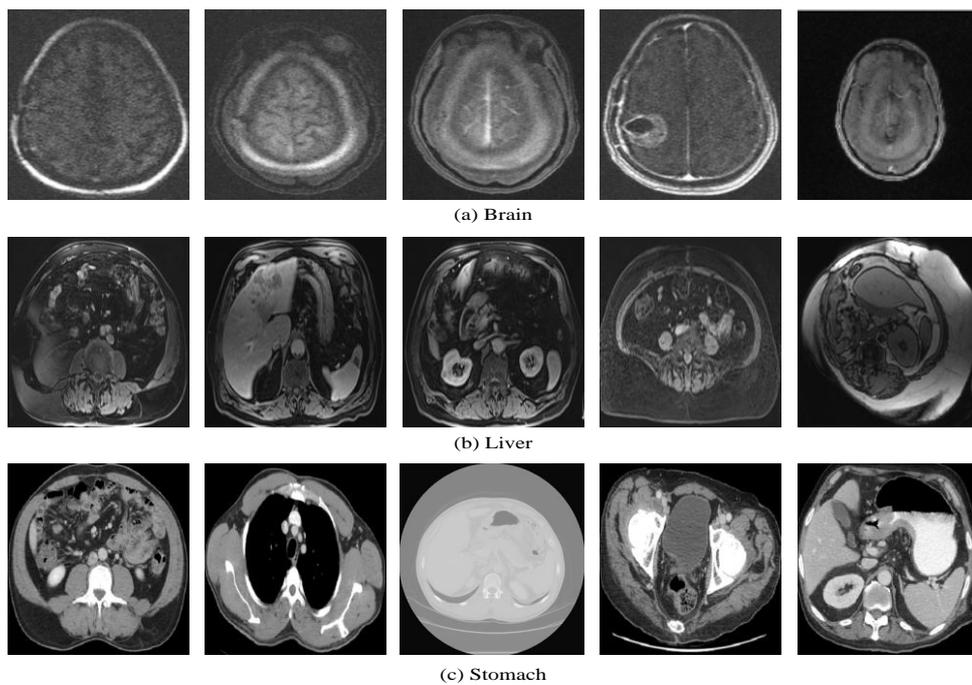

Fig. 6. Example images from different classes showing intra-class variations (a) Brain, (b) Liver and (c) Stomach

## 4.1 Dataset

The dataset used for the proposed CBMIR task has been collected from publicly available medical databases and classes were formed according to the body organ e.g. lungs, brain, liver etc. There is a total of 24 classes in the dataset used in this research, out of which data for 22 classes was acquired from various public databases available at cancer imaging archive[1]. The other two classes contain data from Messidor [46] and an open access website for knee images [47]. A total of 300 images were taken from each class on random giving a dataset of 7200 images. The data from each class was divided randomly into training and testing set using 70% and 30% images for training and testing set. A total of 5040 and 2160 images were used in training and testing set respectively. The training and testing sets did not contain any similar image. All the images were in DICOM (Digital Imaging and COmmunication in Medicine) format except that of the images from Messidor, as it contains images in TIF (Tagged Image File) format. All the images from each class were resized to $256 \times 256$ and color images from Eye class were converted into grayscale. Numeric class labels were assigned to classes for supervised learning. Example images from classes used in the dataset are shown in Fig. 5 and Fig. 6, which emphasize that there is interclass as well intra-class variance among images respectively.

## 4.2 Classification Performance

The performance of the proposed framework for classification task is evaluated in terms of average precision (AP), average recall (AR), accuracy and F1 measure which are calculated as

$$AP = \frac{1}{N}\sum_{i=1}^{N} \frac{TP_i}{TP_i+FP_i}, \qquad (6)$$

$$AR = \frac{1}{N}\sum_{i=1}^{N} \frac{TP_i}{TP_i+TN_i}, \qquad (7)$$

$$Accuracy\ (Acc) = \frac{1}{N}\sum_{i=1}^{N} \frac{TP_i+TN_i}{TP_i+TN_i+FP_i+FN_i}, \qquad (8)$$

$$F1\ measure = 2 \times \frac{AP \times AR}{AP + AR}, \qquad (9)$$

where, TP is true positive and denote the number of images from class $k$ and correctly classified, FP is false positive and denote number of images not from class $k$ but misclassified as class $k$, TN is true negative and denote number of images that are correctly classified as not belonging to class $k$, FN is false negative and denote the images that are from class $k$ but are misclassified and 'N' represents the total number of classes that equals 24 in this case. A 10-fold cross validation was used on the training data and the testing set was tested 100 times giving an AP, AR and average F1 measure of 99.76, 99.77 and 99.76 respectively. The confusion matrix is shown in Table 1, where each class gives an average accuracy of 100 % except stomach, liver and bladder classes that gives an average accuracy of 98.9%, 97.7% and 98.9% respectively. The classification performance is compared with single modality organ classification method [43] and is summarized in Table 2. The proposed system performs better in classifying organs when applied to multimodal data. Although the system trained in [43] is on a different set of image collection, but the high accuracy achieved by our proposed system demonstrates the efficacy of the method in the classification task.

## 4.3 Retrieval Performance

The performance of the proposed framework for CBMIR has been tested using most frequently used performance measure for CBIR systems i.e., Precision and Recall. The mathematical expression for precision and recall are,

$$Precision = \frac{Number\ of\ relevant\ images\ retrieved}{Number\ of\ retrieved\ images}, \qquad (10)$$

---

[1] www.cancerimagingarchive.net

$$Recall = \frac{Number\ of\ relevant\ images\ retrieved}{Total\ Number\ of\ relevant\ images\ in\ database}. \quad (11)$$

Table 1. Confusion matrix of medical image classification with 24 classes using DCNN.

| | Brain | Liver | Stomach | Soft Tissue | Chest | Breast | Renal | Thyroid | Phantom | Rectum | Bladder | Uterus | Head-Neck | Esophagus | Cervix | Prostate | Ovary | Colon | Abdomen | Pancreas | Kidney | Knee | Lungs | Eye |
|---|---|---|---|---|---|---|---|---|---|---|---|---|---|---|---|---|---|---|---|---|---|---|---|---|
| Brain | 100.0 | 0 | 0 | 0 | 0 | 0 | 0 | 0 | 0 | 0 | 0 | 0 | 0 | 0 | 0 | 0 | 0 | 0 | 0 | 0 | 0 | 0 | 0 | 0 |
| Liver | 0 | 98.9 | 0 | 0 | 0 | 0 | 0 | 0 | 0 | 0 | 0 | 0 | 0 | 0 | 1.1 | 0 | 0 | 0 | 0 | 0 | 0 | 0 | 0 | 0 |
| Stomach | 0 | 0 | 96.7 | 0 | 0 | 0 | 0 | 3.3 | 0 | 0 | 0 | 0 | 0 | 0 | 0 | 0 | 0 | 0 | 0 | 0 | 0 | 0 | 0 | 0 |
| Soft Tissue | 0 | 0 | 0 | 100.0 | 0 | 0 | 0 | 0 | 0 | 0 | 0 | 0 | 0 | 0 | 0 | 0 | 0 | 0 | 0 | 0 | 0 | 0 | 0 | 0 |
| Chest | 0 | 0 | 0 | 0 | 100.0 | 0 | 0 | 0 | 0 | 0 | 0 | 0 | 0 | 0 | 0 | 0 | 0 | 0 | 0 | 0 | 0 | 0 | 0 | 0 |
| Breast | 0 | 0 | 0 | 0 | 0 | 100.0 | 0 | 0 | 0 | 0 | 0 | 0 | 0 | 0 | 0 | 0 | 0 | 0 | 0 | 0 | 0 | 0 | 0 | 0 |
| Renal | 0 | 0 | 0 | 0 | 0 | 0 | 100.0 | 0 | 0 | 0 | 0 | 0 | 0 | 0 | 0 | 0 | 0 | 0 | 0 | 0 | 0 | 0 | 0 | 0 |
| Thyroid | 0 | 0 | 0 | 0 | 0 | 0 | 0 | 100.0 | 0 | 0 | 0 | 0 | 0 | 0 | 0 | 0 | 0 | 0 | 0 | 0 | 0 | 0 | 0 | 0 |
| Phantom | 0 | 0 | 0 | 0 | 0 | 0 | 0 | 0 | 100.0 | 0 | 0 | 0 | 0 | 0 | 0 | 0 | 0 | 0 | 0 | 0 | 0 | 0 | 0 | 0 |
| Rectum | 0 | 0 | 0 | 0 | 0 | 0 | 0 | 0 | 0 | 100.0 | 0 | 0 | 0 | 0 | 0 | 0 | 0 | 0 | 0 | 0 | 0 | 0 | 0 | 0 |
| Bladder | 0 | 0 | 0 | 0 | 0 | 0 | 0 | 0 | 0 | 0 | 100.0 | 0 | 0 | 0 | 0 | 0 | 0 | 0 | 0 | 0 | 0 | 0 | 0 | 0 |
| Uterus | 0 | 0 | 0 | 0 | 0 | 0 | 0 | 0 | 0 | 0 | 0 | 98.9 | 0 | 0 | 0 | 0 | 0 | 0 | 0 | 0 | 1.1 | 0 | 0 | 0 |
| Head-Neck | 0 | 0 | 0 | 0 | 0 | 0 | 0 | 0 | 0 | 0 | 0 | 0 | 100.0 | 0 | 0 | 0 | 0 | 0 | 0 | 0 | 0 | 0 | 0 | 0 |
| Esophagus | 0 | 0 | 0 | 0 | 0 | 0 | 0 | 0 | 0 | 0 | 0 | 0 | 0 | 100.0 | 0 | 0 | 0 | 0 | 0 | 0 | 0 | 0 | 0 | 0 |
| Cervix | 0 | 0 | 0 | 0 | 0 | 0 | 0 | 0 | 0 | 0 | 0 | 0 | 0 | 0 | 100.0 | 0 | 0 | 0 | 0 | 0 | 0 | 0 | 0 | 0 |
| Prostate | 0 | 0 | 0 | 0 | 0 | 0 | 0 | 0 | 0 | 0 | 0 | 0 | 0 | 0 | 0 | 100.0 | 0 | 0 | 0 | 0 | 0 | 0 | 0 | 0 |
| Ovary | 0 | 0 | 0 | 0 | 0 | 0 | 0 | 0 | 0 | 0 | 0 | 0 | 0 | 0 | 0 | 0 | 100.0 | 0 | 0 | 0 | 0 | 0 | 0 | 0 |
| Colon | 0 | 0 | 0 | 0 | 0 | 0 | 0 | 0 | 0 | 0 | 0 | 0 | 0 | 0 | 0 | 0 | 0 | 100.0 | 0 | 0 | 0 | 0 | 0 | 0 |
| Abdomen | 0 | 0 | 0 | 0 | 0 | 0 | 0 | 0 | 0 | 0 | 0 | 0 | 0 | 0 | 0 | 0 | 0 | 0 | 100.0 | 0 | 0 | 0 | 0 | 0 |
| Pancreas | 0 | 0 | 0 | 0 | 0 | 0 | 0 | 0 | 0 | 0 | 0 | 0 | 0 | 0 | 0 | 0 | 0 | 0 | 0 | 100.0 | 0 | 0 | 0 | 0 |
| Kidney | 0 | 0 | 0 | 0 | 0 | 0 | 0 | 0 | 0 | 0 | 0 | 0 | 0 | 0 | 0 | 0 | 0 | 0 | 0 | 0 | 100.0 | 0 | 0 | 0 |
| Knee | 0 | 0 | 0 | 0 | 0 | 0 | 0 | 0 | 0 | 0 | 0 | 0 | 0 | 0 | 0 | 0 | 0 | 0 | 0 | 0 | 0 | 100.0 | 0 | 0 |
| Lungs | 0 | 0 | 0 | 0 | 0 | 0 | 0 | 0 | 0 | 0 | 0 | 0 | 0 | 0 | 0 | 0 | 0 | 0 | 0 | 0 | 0 | 0 | 100.0 | 0 |
| Eye | 0 | 0 | 0 | 0 | 0 | 0 | 0 | 0 | 0 | 0 | 0 | 0 | 0 | 0 | 0 | 0 | 0 | 0 | 0 | 0 | 0 | 0 | 0 | 100.0 |

Table 2. Comparison of classification performance of the proposed DCNN with single modality classification algorithm.

| Method | Number of Images | Training (Images) | Testing (Images) | Modalities | Classes | AP (%) | AR (%) | F1 measure |
|---|---|---|---|---|---|---|---|---|
| DCNN trained on whole images (proposed) | 7200 | 5040 | 2160 | MR, CT, PET, PT, OPT | 24 | 99.76 | 99.77 | 99.76 |
| Two stage DCNN trained on local image patches [43] | 7489 | 2413 (With Data Augmentation on extracted patches) | 4043 | CT | 12 | 98.43 | 97.28 | 97.85 |

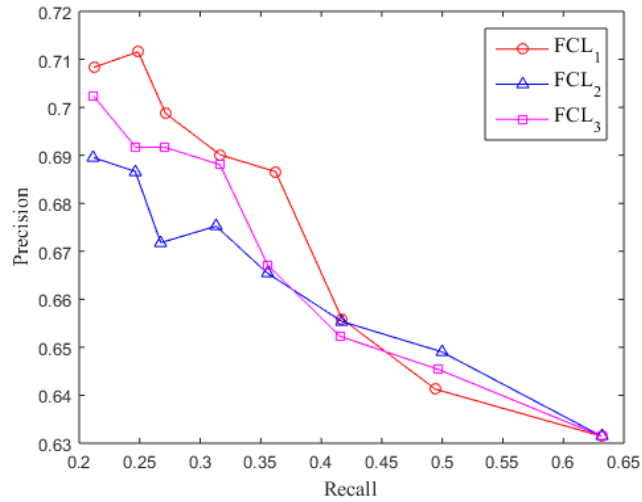

Fig. 7. Precision vs Recall for CBMIR with class prediction.

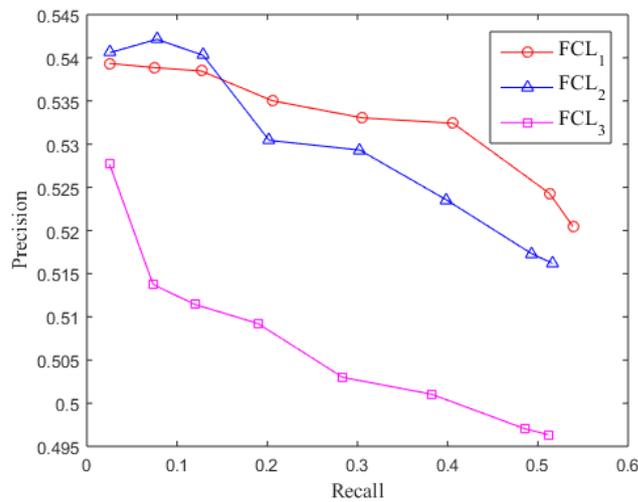

Fig. 8. Precision vs Recall for CBMIR without class prediction.

Feature representations from all three fully connected layers of the trained model have been used for retrieval of medical images. Analysis of these representations has been performed in terms of retrieval quality using both options i.e., with and without using the predicted class labels. The precision vs recall plots for feature representations extracted from $FCL_1$, $FCL_2$, and $FCL_3$ using class prediction and without using class prediction are depicted in Fig. 7 and Fig. 8 respectively. It shows that precision is high for feature representations of $FCL_1$ as compared to feature representations of layer $FCL_2$ and $FCL_3$. The improvement in performance in terms of precision is also evident in case of using class predictions.

The retrieval performance has also been evaluated using mean average precision with and without class predictions. Fig. 9 and 10 shows the retrieved images, of query image from chest and renal classes with and without class predictions respectively. The retrieval results are shown in a ranked order, where the most relevant image found after feature comparison is presented first. The retrieved results demonstrate the interclass variance. When class predictions are not used images from other classes are also retrieved highlighted by red boxes in Fig. 9 and Fig. 10.

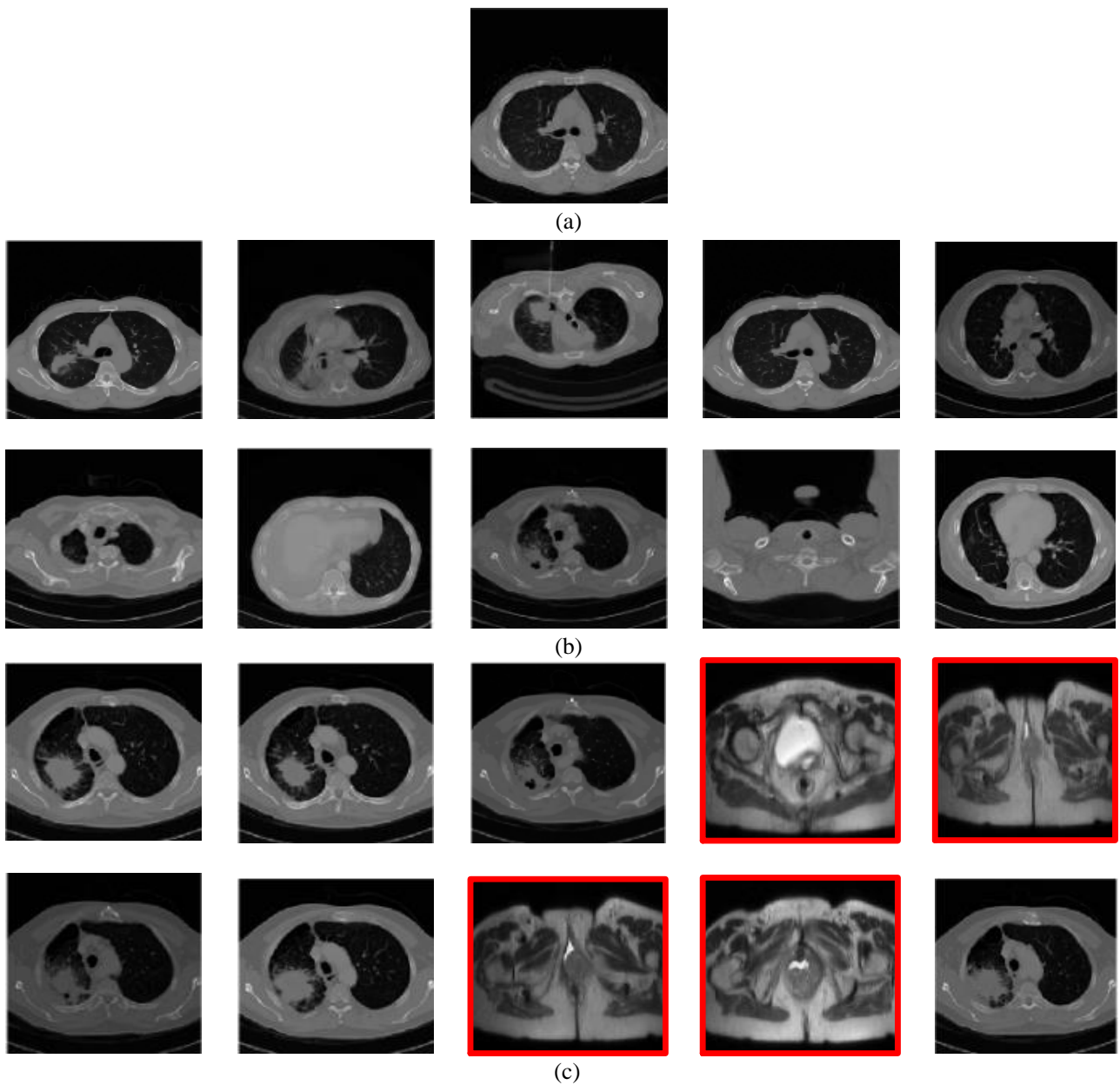

Fig. 9. Retrieval results for chest class (a) query image (b) retrieved images using class prediction (c) retrieved images without using class prediction.

The system achieves a mean average precision of 0.53 and 0.69 without class predictions and with class prediction respectively. The retrieval results are improved since the images retrieved only belong to the predicted class by the classification framework.

### 4.4 Comparison

To evaluate our proposed deep learning based framework for medical image retrieval, a comparison is made with some recent systems used for such task. As direct comparison was not possible, since to the best of our knowledge, no standard medical dataset is available that can be used to benchmark the retrieval system. Hence, two criterions have been used to make the comparison, one is classification accuracy, average precision, and average recall for classification (Table 2) and the other is mean average precision (mAP) for retrieval given in Table 3. Although, [48] achieves a higher value for mAP, but they have only worked for a single modality whereas, our proposed system work for multimodal data.

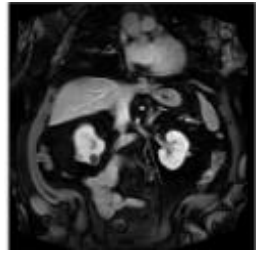

(a)

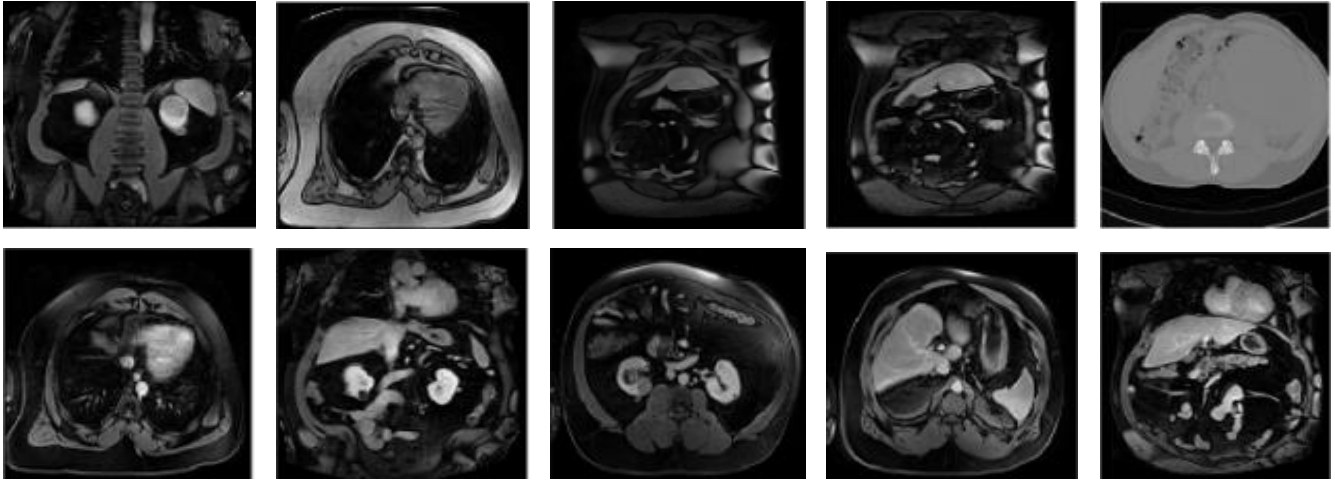

(b)

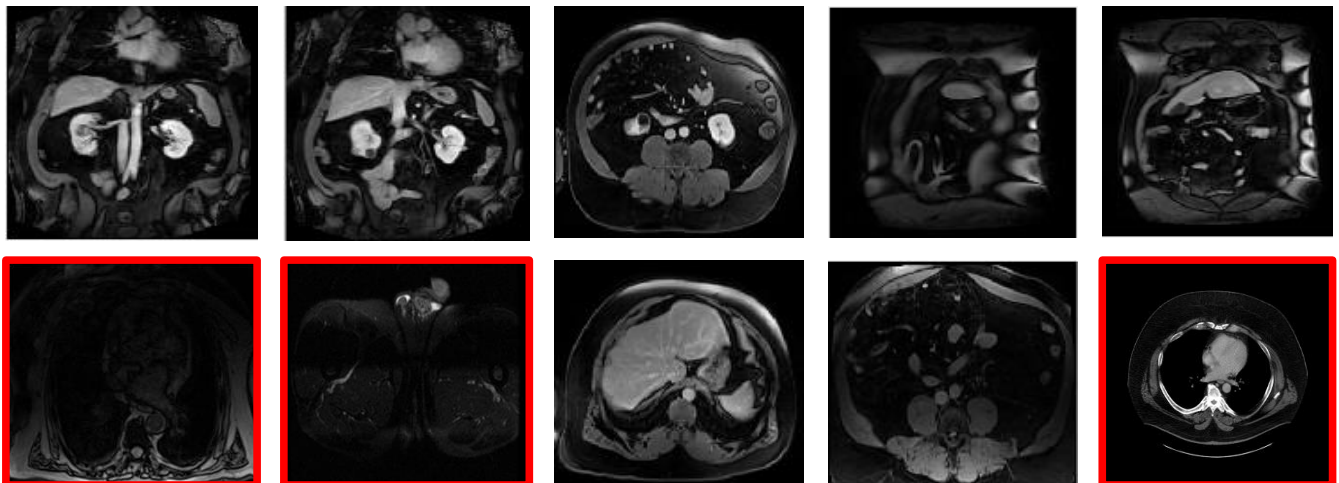

(c)

Fig. 10. Retrieval results for renal class (a) query images (b) retrieved images using class prediction c) retrieved images without using class prediction.

Table 3. Comparison of the proposed CBMIR using deep learning with state of the art systems in terms of mean average precision

| Method | Images | Training (Images) | Testing (Images) | Modalities | Classes | Mean Average Precision (mAP) |
|---|---|---|---|---|---|---|
| DCNN trained on whole images (proposed) | 7200 | 5040 | 2160 | MR, CT, PT, PET, OPT | 24 | 0.53 without using class predictions<br><br>0.69 using class predictions |
| Local Binary Patterns, Support Vector Machine, and Auto Encoder [48] | 14410 | 12677 | 1733 | X-Ray | 57 | 0.86 |
| Full Robe Auto Regressive Model and Binary Tree Based SVM [49] | 6400 | 5760 | 640 | CT, MRI, Mammography, Microscopy, Ultrasound, Endoscopy | 83 | 0.576 |

## 5. Conclusion

This paper proposes a deep learning based framework for content based medical image retrieval by training a deep convolutional neural network for the classification task. Two strategies have been proposed for retrieval of medical images, one is by getting prediction about the class of query image by the trained network and then to search relevant images in that specific class. The second method is without incorporating the information about the class of the query image and therefore searching the whole database for relevant images. The proposed solution reduces the semantic gap by learning discriminative features directly from the images. The network was successfully trained for 24 classes of medical images with an average classification accuracy of 99.77%. The last three fully connected layers of the network have been used to extract features for the retrieval task. Widely used metrics i.e., precision and recall were used to test the performance of the proposed framework for medical image retrieval. The proposed system achieves a mean average precision of 0.69 for multimodal image data with class prediction. We intend to further improve the retrieval performance by using a larger dataset and adapt the network for 3D volumetric applications by defining further classes that incorporate the different geometric views of a 3D slice.